\documentclass{svmult}

\usepackage{mathptmx}       
\usepackage{helvet}         
\usepackage{courier}        
\usepackage{type1cm}        
\usepackage{graphicx}        
\usepackage{multicol}        
\usepackage[bottom]{footmisc}
\usepackage{amsmath,amssymb,epsfig}

\begin{document}

\title*{Stiffness modelling of parallelogram-based parallel manipulators}
\author{A. Pashkevich, A. Klimchik, S. Caro and D. Chablat}
\institute{Institut de Recherche en Communications et en Cybernetique de Nantes, France  \\
Ecole des Mines de Nantes, France  \\
\email{ anatol.pashkevich@emn.fr, alexandr.klimchik@emn.fr, stephane.caro@irccyn.ec-nantes.fr, damien.chablat@irccyn.ec-nantes.fr}}

%
%
%
\maketitle

\abstract{The paper presents a methodology to enhance the stiffness analysis of parallel manipulators with parallelogram-based linkage. It directly takes into account the influence of the external loading and allows computing both the non-linear ``load-deflection" relation and relevant rank-deficient stiffness matrix. An equivalent bar-type pseudo-rigid model is also proposed to describe the parallelogram stiffness by means of five mutually coupled virtual springs. The contributions of this paper are highlighted with a parallelogram-type linkage used in a manipulator from the Orthoglide family.}

\keywords{stiffness modeling, parallel manipulators, parallelogram-based linkage, external loading, linear approximation.}

\section{Introduction}

In the last decades, parallel manipulators have received growing attention in industrial robotics due to their inherent advantages of providing better accuracy, lower mass/inertia properties, and higher structural rigidity compared to their serial counterparts (Ceccarelli et al., 2002, Company et al., 2002, Merlet, 2000). These features are induced by the specific kinematic structure, which eliminates the cantilever-type loading and decreases deflections due to external wrehches. Accordingly, they are used in innovative robotic systems, but practical utilization of the potential benefits requires development of efficient stiffness modeling techniques, which satisfy the computational speed and accuracy requirements of relevant design procedures.

Amongst numerous parallel architectures studied in robotics literature, of special interest are the parallelogram-based manipulators that employ special type of linkage constraining undesirable motions of the end-platform. However, relevant stiffness analysis is quite complicated due to the presence of internal closed kinematic chains that are usually replaced by equivalent limbs (Majou et al., 2007, Pashkevich et al., 2009a) whose parameters are evaluated rather intuitively. Thus, the problem of adequate stiffness modelling of parallelogram-type linkages, which is in the focus of this paper, is still a challenge and requires some developments.          

Another important research issue is related to the influence of the external loading that may change the stiffness properties of the manipulator. In this case, in addition to the conventional ``elastic stiffness" in the joints, it is necessary to take into account the ``geometrical stiffness" due to the change in the manipulator configuration under the load (Kovecses and Anjeles, 2007). Moreover buckling phenomena may appear (Timoshenko and Goodier, 1970) and produce structural failures, which must be detected by relevant stiffness models.

This paper is based on our previous work on the stiffness analysis of over-constrained parallel architectures (Pashkevich et al., 2009a, 2009b) and presents new results by considering the loading influence on the manipulator configuration and, consequently, on its Jacobian and Hessian. It implements the virtual joint method (VJM) introduce by Salisbary (Salisbury, 1980) and Gosselin (Gosselin, 1990) that describes the manipulator element compliance with a set of localized six-dimensional springs separated by rigid links and perfect joints. The main contribution of the paper is the introduction of a non-linear stiffness model of the parallelogram-type linkage and its linear approximation by 6x6 matrix of the rank 5, for which it is derived an analytical expression.

\section{Problem statement}

\begin{figure}
\center
\includegraphics[width=11.5cm]{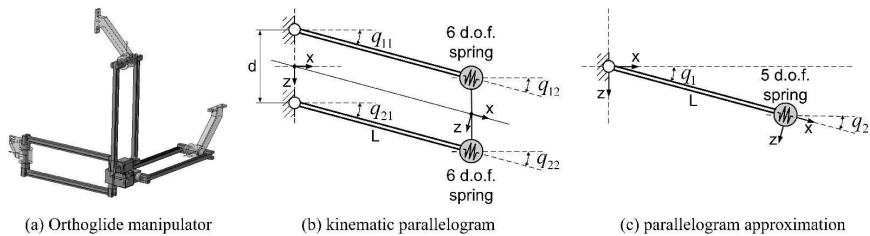}
\caption{Parallel manipulators with parallelogram-based linkages.}
\label{Figure:1}
\end{figure}

Let us consider a typical parallel manipulator that is composed of several kinematic chains connecting a fixed base and a moving platform (Figure~\ref{Figure:1}a). It is assumed, that at least one chain includes a parallelogram-based linkage that may introduce some redundant constraints improving the mechanism stiffness. Following the VJM concept, the manipulator chains are usually presented as a serial sequence of pseudo-rigid links separated by rotational or translational joints of one of the following types: (i) perfect passive joints; (ii) perfect actuated joints, and (iii) virtual flexible joints that describe compliance of the links, joints and/or actuators.

To derive a parallel architecture without internal kinematic loops, the parallelogram based linkage must be replaced by an equivalent bar-type element that has similar elastic properties. Thus, let us consider the VJM model of the kinematic parallelogram assuming its compliance is mainly due to the compliance of the longest links, which are oriented in the direction of the linkage (Figure~\ref{Figure:1}b,c). In order to be precise, let us also assume that the stiffness of the original bar-elements is described by a $6 \times 6$ matrix $\mathbf{K}_{b}$ whose elements are identified using FEA-based modeling. 

In the framework of the VJM approach, the geometry of the parallelogram-based linkage can be descried by the following homogeneous transformations

\begin{equation}\label{Eq:1}
\begin{array}{l}

 \mathbf{T}_i = \mathbf{T}_z(\eta~d/2)~   \mathbf{R}_y(q_{i1})~  \mathbf{T}_x(L)~   \mathbf{T}_x(\theta_{i1})~   \mathbf{T}_y(\theta _{i2})~   \mathbf{T}_z(\theta_{i3})~   \mathbf{R}_x(\theta_{i4})
 \\[3pt]
 ~~~~~~~~~~~~~~~~~~~~~~~~~~~~~~~~~~~~~~~~\mathbf{R}_y(\theta_{i5})~   \mathbf{R}_z(\theta_{i6})~   \mathbf{R}_y(q_{i2})~   \mathbf{T}_z(-\eta~d/2)~   \mathbf{R}_y(-q_{i2}) \\ 
 \end{array}
\end{equation}
where $\eta=(-1)^{i}$, $\mathbf{T}_{i}$  is a $4 \times 4$ homogenous transformation matrix, $\mathbf{T}_{x}$, ..., $\mathbf{R}_{z}$ are the matrices of elementary translation and rotation, $L$, $d$ are the parallelogram geometrical parameters, $q_{ij}$  are the passive joint coordinates, $\theta_{ij}$  are the virtual joint coordinates, $i=1,2$ defines the number of the kinematic chain, $j$  identifies the coordinate number within the chain. It should be noted that using this notation, the closed loop equation can be expressed as $\mathbf{T}_{1}=\mathbf{T}_{2}$ . 

For further computational convenience, the homogenous matrix equations (1) may be also rewritten in the vector form as

\begin{equation}\label{Eq:2}
\mathbf{t}=\mathbf{g}_{i}(\mathbf{q}_{i},\mathbf{\theta}_{i}),
\end{equation}
where vector $\mathbf{t}=(\mathbf{p},\mathbf{\varphi})$ is the output frame pose, that includes its position $\mathbf{p}=(x,y,z)^{T}$ and orientation $\mathbf{\varphi}=(\varphi_{x},\varphi_{y},\varphi_{z})$, vector $\mathbf{q}_{i}=(q_{i1},q_{i2})$  contains the passive joint coordinates, and vector $\mathbf{\theta}_{i}=(\theta_{i1},...,\theta_{i6})$ collects all virtual joint coordinates of the corresponding chain. Using the above assumptions and definitions, let us derive the stiffness model of the parallelogram-based linkage.

\section{Static equilibrium}

Let us derive first the general static equilibrium equations of the parallelogram assuming that it is virtually divided into two symmetrical serial kinematic chains with the geometrical and elastostatic models $\mathbf{t}=\mathbf{g}_{i}(\mathbf{q}_{i},\mathbf{\theta}_{i})$, $\mathbf{\tau}_{i}=\mathbf{K}_{i}~\mathbf{\theta}_{i}$, $i=1,2$ where the variables $\mathbf{\theta}_{i}$ correspond to the deformations in the virtual springs, $\mathbf{K}_{i}$ is the spring stiffness matrix, and $\mathbf{\tau}_{i}$ incorporates corresponding forces and torques. Taking into account that the considered mechanism is under-constrained, it is prudent to assume that the end-points of both chains are located at a given position $\mathbf{t}$ and to compute partial forces $\mathbf{F}_{i}$ corresponding to the static equilibrium. 

To derive equations, let us express the potential energy of the mechanism as  

\begin{equation}\label{Eq:3}
E(\mathbf{\theta}_{1},\mathbf{\theta}_{2})=\frac{1}{2}~\sum\limits_{i=1}^{2}\mathbf{\theta}_{i}^{T}~\mathbf{K}_{\theta i}~\mathbf{\theta}_{i}
\end{equation}
In the equilibrium configuration, this energy must be minimised subject to the geometrical constraints $\mathbf{t}=\mathbf{g}_{i}(\mathbf{q}_{i},\mathbf{\theta}_{i})$, $i=1,2$. Hence, the Lagrange function is 

\begin{equation}\label{Eq:4}
L(\mathbf{\theta}_{1},\mathbf{\theta}_{2},\mathbf{q}_{1},\mathbf{q}_{2})=\frac{1}{2}~\sum\limits_{i=1}^{2}{\mathbf{\theta }_{i}^{T}~\mathbf{K}_{\theta i}~\mathbf{\theta}_{i}}+\sum\limits_{i=1}^{2}{\mathbf{\lambda}_{i}^{T}}\left( \mathbf{t}-\mathbf{g}_{i}(\mathbf{\theta}_{i},\mathbf{q}_{i}) \right)
\end{equation}
where the multipliers $\mathbf{\lambda}_{i}$ may be interpreted as the external forces $\mathbf{F}_{i}$ applied at the end-points of the chains. Further, after computing the partial derivatives of $L(...)$ with respect to $\mathbf{\theta}_{i},\mathbf{q}_{i},\mathbf{\lambda}_{i}$ and setting the derivatives to zero, equations of the static equilibrium can be presented as

\begin{equation}\label{Eq:5}
\mathbf{J}_{\theta i}^{T}\mathbf{\lambda}_{i}=\mathbf{K}_{i}~\mathbf{\theta}_{i};~~~~~~~~~~  
\mathbf{J}_{qi}^{T}~\mathbf{\lambda}_{i}=\mathbf{0};~~~~~~~~~~
\mathbf{t}=\mathbf{g}_{i}(\mathbf{q}_{i},\mathbf{\theta}_{i});~~~~~~~~~~
i=1,2
\end{equation}
where $\mathbf{J}_{\theta i}={\partial \mathbf{g}_{i}(...)}/\partial {\mathbf{\theta}_{i}}$ and $\mathbf{J}_{qi}=\partial \mathbf{g}_{i}(...)/\partial {\mathbf{q}_{i}}$ are the kinematic Jacobians. Here, the configuration variables $\mathbf{\theta}_{i},~\mathbf{q}_{i}$ and the multipliers $\mathbf{\lambda}_{i}$ are treated as unknowns.

Since the derived system is highly nonlinear, the desired solution can be obtained only numerically. In this paper, it is proposed to use the following iterative procedure

\begin{equation}\label{Eq:6}
\begin{array}{l}
\left[ 	
\begin{array}{c}
   \mathbf{\lambda}_{i}'  \\ [3pt]
   \mathbf{{q}'}_\mathbf{i}  \\
\end{array} 
\right]=\left[ 	
\begin{array}{*{20}{c}}
   {\mathbf{J}_{\theta i}~\mathbf{K}_{\theta i}^{-1}~\mathbf{J}_{\theta i}^{T}~~~} & {\mathbf{J}_{qi}}  \\ [3pt]
   {\mathbf{J}_{qi}^{T}~~~} & {\mathbf{0}}  \\
\end{array} 
\right]~ \left[ 
\begin{array}{c}
   \mathbf{t}-\mathbf{g}_{i}+\mathbf{J}_{qi}~\mathbf{q}_{i}+\mathbf{J}_{\theta i}~\mathbf{\theta}_{i}  \\ [3pt]
   \mathbf{0}  \\
\end{array}  \right];~~~
\mathbf{\theta}_{i}'=\mathbf{K}_{\theta i}^{-1}~\mathbf{J}_{\theta i}^{T}~\mathbf{{\lambda}'}_{i}
\end{array} 
\end{equation}
where the symbol `` $'$ "  corresponds to the next iteration. Using this iterative procedure, for any given location of the end-point $\mathbf{t}$, one can compute both the partial forces $\mathbf{F}_{i}$ and the total external force, which allows to obtain the desired ``force-deflection" relation $\mathbf{F}=\mathbf{f}(\mathbf{t})$ for any initial unloaded configuration.

\section{Stiffness matrix}

To compute the stiffness matrix of the considered parallelogram-based mechanism, the obtained ``force-deflection" relations must be linearized in the neighborhood of the static equilibrium. Let assume that the external forces $\mathbf{F}_{i}$, $i=1,2$ and the end-point location $\mathbf{t}$ of both kinematic chains are incremented by some small values $\delta {{\mathbf{F}}_{i}}$, $\delta \mathbf{t}$ and consider simultaneously two equilibriums corresponding to the state variables $(\mathbf{F}_{i},\mathbf{\theta}_{i},\mathbf{q}_{i},\mathbf{t})$ and $(\mathbf{F}_{i}+\delta \mathbf{F}_{i},\mathbf{\theta}_{i}+\delta \mathbf{\theta}_{i},\mathbf{q}_{i}+\delta \mathbf{q}_{i},\mathbf{t}+\delta \mathbf{t})$. Under these assumptions, the original system (\ref{Eq:5}) should be supplemented by the equations

\begin{equation}\label{Eq:7}
\begin{array}{l}
  \left(\mathbf{J}_{\theta i} + \delta \mathbf{J}_{\theta i} \right)   ^{T}~   \left( \mathbf{\lambda}_{i} + \delta \mathbf{\lambda}_{i} \right) = \mathbf{K}_{\theta i}~   \left(\mathbf{\theta}_{i}+\delta \mathbf{\theta}_{i} \right)  ;\\[5pt]
  \left( \mathbf{J}_{qi}+\delta \mathbf{J}_{qi} \right)~  \left( \mathbf{\lambda}_{i}+\delta \mathbf{\lambda}_{i} \right)  = \mathbf{0}; \\ [5pt] 
  \mathbf{t} + \delta \mathbf{t} = \mathbf{g}_{i} (\mathbf{q}_{i},\mathbf{\theta}_{i}) + \mathbf{J}_{\theta i}~\delta \mathbf{\theta}_{i} + \mathbf{J}_{qi}~\delta \mathbf{q}_{i}; \\   
  \end{array}
\end{equation}
where $\delta \mathbf{J}_{\theta i}$ and $\delta \mathbf{J}_{qi}$ are the differentials of the Jacobians due to changes in $(\mathbf{\theta}_{i},\mathbf{q}_{i})$. 
After relevant transformation, neglecting high-order small terms and expanding the differentials via the Hessians of the function $\Psi_{i}=\mathbf{g}_{i} (\mathbf{q}_{i},\mathbf{\theta}_{i})^{T}~ \mathbf{\lambda}_{i}$ : $\mathbf{H}_{qq}^{(i)}=\partial ^{2} \Psi_{i} / \partial \mathbf{q}_{i}^{2}$, $\mathbf{H}_{q \theta}^{(i)}=\partial ^{2} \Psi_{i} / \partial \mathbf{q}_{i}^ \partial \mathbf{\theta}_{i}$, $\mathbf{H}_{\theta \theta }^{(i)}= \partial ^{2} \Psi_{i} / \partial \mathbf{\theta }_{i}^{2}$, the system of equations may be rewritten as

\begin{equation}\label{Eq:9}
\begin{array}{l}
	\mathbf{J}_{\theta i}~ \delta \mathbf{\lambda}_{i} + \mathbf{H}_{q\theta}^{(i)}~ \delta \mathbf{q}_{i} + \mathbf{H}_{\theta \theta}^{(i)}~ \delta \mathbf{\theta}_{i}=\mathbf{K}_{\theta i}~ \delta \mathbf{\theta}_{i}; \\ [3pt]
  \mathbf{J}_{qi}~ \delta \mathbf{\lambda}_{i}+\mathbf{H}_{qq}^{(i)}~ \delta \mathbf{q}_{i}+\mathbf{H}_{q\theta}^{(i)}~ \delta \mathbf{\theta}_{i}=\mathbf{0};\\ [3pt]
  \mathbf{J}_{\theta i}~ \delta \mathbf{\theta}_{i} + \mathbf{J}_{qi}~ \delta \mathbf{q}_{i}=\delta \mathbf{t}
  \end{array}
\end{equation}

After analytical elimination of the variable $\delta \mathbf{\theta}_{i}$ and defining $\mathbf{k}_{\theta}^{(i)}=(\mathbf{K}_{\theta}-\mathbf{H}_{\theta \theta}^{(i)})^{-1}$, one can obtain a matrix equation

\begin{equation}\label{Eq:10}
\left[ \begin{array} {c}
   \delta \mathbf{\lambda}_{i}  \\ [3pt]
   \delta \mathbf{q}_{i}  \\
\end{array} \right]=
\left[ {\begin{array}{*{20}{c}}
   \mathbf{J}_{\theta i} \cdot \mathbf{k}_{\theta}^{(i)} \cdot \mathbf{J}_{\theta i}^{T} & \;\;
   \mathbf{J}_{q i} + \mathbf{J}_{\theta i} \cdot \mathbf{k}_{\theta}^{(i)} \cdot \mathbf{H}_{\theta q}^{(i)}  \\ [3pt]
   \mathbf{J}_{q i}^{T} + \mathbf{H}_{q\theta }^{(i)} \cdot \mathbf{k}_{\theta}^{(i)} \cdot \mathbf{J}_{\theta i}^{T} & \;\;
   \mathbf{H}_{qq}^{(i)} + \mathbf{H}_{q\theta }^{(i)} \cdot \mathbf{k}_{\theta}^{(i)} \cdot \mathbf{H}_{\theta q}^{(i)}  \\
\end{array}} \right]^{-1} \cdot
\left[ \begin{array}{c} \delta \mathbf{t}  \\ [3pt]
   \mathbf{0}  \\
\end{array} \right]\\ 
\end{equation}
which yields the linear relation $\delta \mathbf{\lambda}=\mathbf{K}_{ci}~ \delta \mathbf{t}$ defining the Cartesian stiffness matrix $\mathbf{K}_{ci}$ for each kinematic chain. Taking into account the architecture of the considered mechanism, the total stiffness matrix may be found as  $\mathbf{K}_{c}=\mathbf{K}_{c1}+\mathbf{K}_{c2}$. These expressions allow numerical computation of the Cartesian stiffness matrix for a general case, both for loaded and unloaded equilibrium configuration.

However, in the case of unloaded equilibrium, the above presented equations may be essentially simplified:

\begin{equation}\label{Eq:11} 
\mathbf{K}_{p}(q) = 2 \left[\small \begin{array}{ccc|ccc}
K_{11} & 0      & ~~0~ & ~0 & 0 & 0\\ [3pt]
0      & K_{22} & ~~0~ & ~0 & 0 & K_{26}\\ [3pt] 
0      & 0      & ~~0~ & ~0 & 0 & 0\\ [3pt] \hline 
0      & 0      & ~~0~ & ~K_{44}+\dfrac{d^{2}~C^{2}_{q}~K_{22}}{4} & 0 & \dfrac{d^{2}~S_{2q}~K_{22}}{8}\\ [3pt]
0      & 0      & ~~0~ & ~0 & \dfrac{d^{2}~C^{2}_{q}~K_{11}}{4} & 0\\ [3pt]
0      & K_{26} & ~~0~ & ~\dfrac{d^{2}~S_{2q}~K_{22}}{8} & 0 & K_{66}+\dfrac{d^{2}~S^{2}_{q}~K_{22}}{4}\\ 
\end{array}
\right],
\end{equation}
where $C_{q}=\cos q$, $S_{q}=\sin q$, $S_{2q}=\sin 2q$, $q={{q}_{1i}}$ (Figure 1b,c), $K_{ij}$ are the elements  of the $6 \times 6$ stiffness matrix of the parallelogram bars $\mathbf{K}_{b}$ which are assumed here to be the only elements of the linkage that posses non-negligible stiffness. In the following example, this expression is evaluated from point of view of accuracy and applicability to stiffness analysis of the parallelogram-based manipulators.

\section{Illustrative example}

To demonstrate validity of the proposed model and to evaluate its applicability range, let us apply it to the stiffness analysis of the Orthoglide manipulator (Figure~1a). It includes three parallelogram-type linkage where the main flexibility source is concentrated in the bar elements of length 310 mm. Using the FEA-based software tools and dedicated identification technique (Pashkevich et al., 2009c), the stiffness matrix of the bar element was evaluated as

\begin{equation}\label{Eq:12} 
\mathbf{K}_{b} = \left[\small{\begin{array}{ccc|ccc}
2.20\cdot10^{4} & 0      & 0 & 0 & 0 & 0\\ [3pt]
0      & 1.81\cdot10^{1} & 0 & 0 & 0 & -2.84\cdot10^{3}\\ [3pt] 
0      & 0     & 7.86\cdot10^{1} & 0 & 1.25\cdot10^{4} & 0\\ [3pt] \hline
0      & 0     & 0 & ~~3.48\cdot10^{4} & 0 & 0\\ [3pt]
0      & 0     & 1.25\cdot10^{4}~ & 0 & 2.66\cdot10^{6} & 0\\ [3pt]
0      & -2.84\cdot10^{3} & 0 & 0 & 0 & 5.85\cdot10^{5}\\ \end{array}}
\right],
\end{equation}
where for linear/angular displacements and for the force/torque are used the following units: [mm], [rad], [N], [N·mm]. This bar-element was also evaluated with respect to the structural stability under compression, and the FEA-modelling produced three possible buckling configurations presented in Figure 2. It is obvious that these configurations are potentially dangerous for the compressed parallelogram. However, the parallelogram may also produce some other types of buckling because of presents of the passive joints.

\begin{figure}
\center
\includegraphics[width=11.5cm]{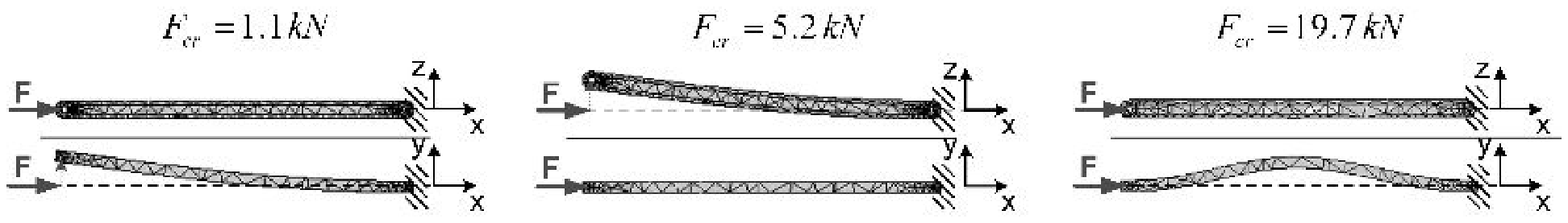}
\caption{Critical forces and buckling configurations of a bar element }
employed in the parallelograms of Orthoglide.
\label{Figure:2}
\end{figure}

For the parallelogram-base linkage incorporating the above bar elements, expression (\ref{Eq:12}) yields the following stiffness matrix 

\begin{equation}\label{Eq:13} 
\mathbf{K}_{p} = 2\left[\small{\begin{array}{ccc|ccc}
2.20\cdot10^{4} & 0      & ~~0~~ & 0 & 0 & 0\\ [3pt]
0      & 1.81\cdot10^{1} & ~~0~~ & 0 & 0 & -2.84\cdot10^{3}\\ [3pt] 
0      & 0     & ~~0~~ & 0 & 0 & 0\\ [3pt] \hline{.} 
0      & 0     & ~~0~~ & ~~5.64\cdot10^{4} & 0 & 1.25\cdot10^{4}\\ [3pt]
0      & 0     & ~~0~~ & 0 & 2.64\cdot10^{7} & 0\\ [3pt]
0      & -2.84\cdot10^{3} & ~~0~~ & ~~1.25\cdot10^{4} & 0 & 5.92\cdot10^{5}\\ 
\end{array}}
\right],
\end{equation}
corresponding to the straight configuration of the linkage (i.e. to $q=0$). Being in good agreement with physical sense, this matrix is rank deficient and incorporates exactly one zero row/column corresponding to the z-translation, where the parallelogram is completely non-resistant due to specific arrangement of passive joints. Also, as follows from comparison with doubled stiffness matrix (\ref{Eq:12}) that may be used as a reference, the parallelogram demonstrates essentially higher rotational stiffness that mainly depends on the translational stiffness parameters of the bar (moreover, the rotational stiffness parameter $K_{55}$ of the bar is completely eliminated by the passive joints). The most significant here is the parallelogram width $d$ that explicitly presented in the rotational sub-block of $\mathbf{K}_{p}$. 

To investigate applicability range of the linear model based on the stiffness matrix (\ref{Eq:11}), it was computed a non-linear ``force-deflection" relation corresponding to the parallelogram compression in the x-direction. This computation was performed using an iterative algorithm presented in Section 3. As follows from obtained results, the matrix (\ref{Eq:11}) ensures rather accurate description of the parallelogram stiffness in small-deflection area. However, for large deflections, corresponding VJM-model detects a geometrical buckling that limits applicability of the matrix (\ref{Eq:11}). 

Similar analysis was performed using the FEA-technique, which yielded almost the same ``force-deflection" plot for the small deflections but detected several types of the buckling, with the critical forces may be both lower and higher then in the VJM-modelling. Corresponding numerical values and parallelogram configurations are presented in Figure 3.

\begin{figure} [b]
\center
\includegraphics[width=10.0cm]{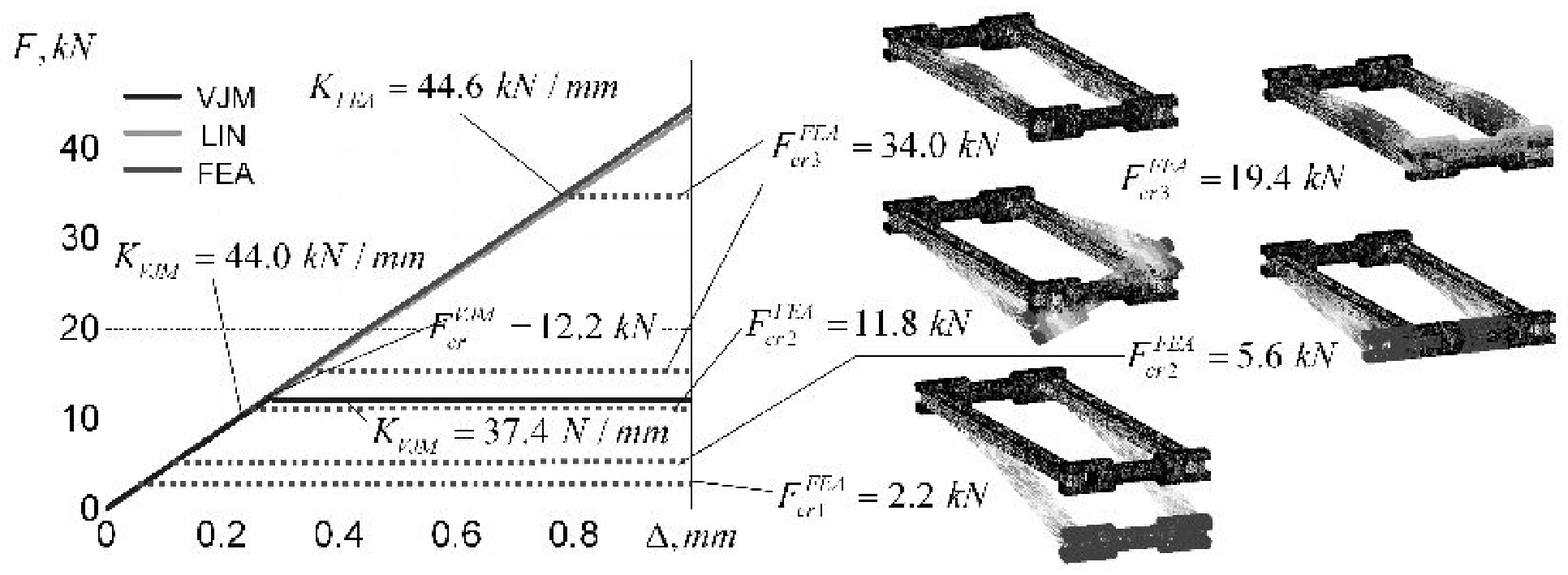}
\caption{Force-deflection relations and buckling configurations for parallelogram compression} 
(modelling methods: VJM, FEA, and LIN - linear model with proposed stiffness matrix).
\label{Figure:3}
\end{figure}

\begin{table} [b]
\caption{Stiffness parameters of the Orthoglide manipulator for different assumptions concerning parallelogram linkage.}
\label{Table:Orthoglide}
\begin{tabular}{p{3.8cm}p{1.8cm}p{1.8cm}p{1.8cm}p{1.8cm}}
\noalign{\smallskip}\svhline\noalign{\smallskip}
Model of linkage & \multicolumn{2} {c} {VJM-model} & \multicolumn{2} {c} {FEA-model}  \\ \cline{2-5}
                                        & $K_{tran}$         & $K_{rot}$          & $K_{tran}           $ & $K_{rot}$        \\
                                        & $[mm/N]$           & $[rad/N]$          & $[mm/N]$            & $[rad/N]$          \\
\hline\noalign{\smallskip} 
2x-bar linkage                          &  $3.35\cdot10^{3}$ & $0.13\cdot10^{6}$  & ---                 & ---                \\
Parallelogram linkage \\(solid axis)    &  $3.23\cdot10^{3}$ & $4.33\cdot10^{6}$  & $3.33\cdot10^{3}$   & $4.10\cdot10^{6}$  \\
Parallelogram linkage \\(flexible axis) &  $3.08\cdot10^{3}$ & $4.06\cdot10^{6}$  & $3.31\cdot10^{3}$   & $4.07\cdot10^{6}$  \\
\noalign{\smallskip}\svhline\noalign{\smallskip}
\end{tabular}
\end{table}

The developed VJM-model of parallelogram was verified in the frame of the stiffness modelling of the entire manipulator. Relevant results are presented in Table~1. They confirm advantages of the parallelogram-based architectures with respect to the translational stiffness and perfectly match to the values obtained from FEA-method. However, releasing some assumptions concerning the stiffness properties of the remaining parallelogram elements (other than bars) modifies the values of the translational and rotational stiffness. The latter gives a new prospective research direction that targeted at more general stiffness model of the parallelogram.

\section{Conclusions}

The paper presents new results in the area of enhanced stiffness modeling of parallel manipulators with parallelogram-based linkage. In contrast to other works, it explicitly takes into account influence of the loading and allows both computing the non-linear load-deflection relation and detecting the buckling phenomena that may lead to manipulator structural failure. There is also derived an analytical expression for relevant rank-deficient stiffness matrix and proposed equivalent bar-type pseudo-rigid model that describes the parallelogram stiffness by five localized mutually coupled virtual springs. These results are validated for a case study that deals with stiffness modeling of a parallel manipulator of the Orthoglide family, for which the parallelogram-type linkage was evaluated using both proposed technique and straightforward FEA-modeling.

\begin{acknowledgement}
The work presented in this paper was partially funded by the Region ``Pays de la Loire" (project RoboComposite).
\end{acknowledgement}



\small
Ceccarelli M., Carbone G., 2002, A stiffness analysis for CaPaMan (Cassino Parallel Manipulator), \textit{Mechanism and Machine Theory} \textbf{37(5):}427-439. 

Company O., Krut S., Pierrot F., 2002, Modelling and preliminary design issues of a 4-axis parallel machine for heavy parts handling. \textit{Journal of Multibody Dynamics} \textbf{216:}1-11.

Gosselin C., 1990, Stiffness mapping for parallel manipulators, \textit{IEEE Transactions on Robotics and Automation} \textbf{6(3):}377-382.

Kovecses J., Angeles J., 2007, The stiffness matrix in elastically articulated rigid-body systems, \textit{Multibody System Dynamics} \textbf{18(2):}169-184.

Majou F., Gosselin C., Wenger P., Chablat D., 2007, Parametric stiffness analysis of the Orthoglide, \textit{Mechanism and Machine Theory} \textbf{42:}296-311.

Merlet, J.-P., 2000, \textit{Parallel Robots,}, Kluwer Academic Publishers, Dordrecht.

Pashkevich A., Chablat D., Wenger Ph., 2009a, Stiffness analysis of overconstrained parallel manipulators, \textit{Mechanism and Machine Theory} \textbf{44:}966-982.

Pashkevich	A., Klimchik A., Chablat D., Wenger Ph., 2009b, Stiffness analysis of multichain parallel robotic systems with loading, \textit{Journal of Automation, Mobile Robotics \& Intelligent Systems} \textbf{3(3):}75-82. 

Pashkevich	A., Klimchik A., Chablat D., Wenger Ph., 2009c, Accuracy improvement for stiffness modeling of parallel manipulators, in \textit{Proceedings of 42nd CIRP Conference on Manufacturing Systems}, Grenoble, France.

Salisbury J., 1980, Active stiffness control of a manipulator in Cartesian coordinates, in \textit{19th IEEE Conference on Decision and Control} 87-97.

Timoshenko S., Goodier J. N, 1970, \textit{Theory of Elasticity,} 3d ed., McGraw-Hill, New York.

\end{document}